\documentclass[runningheads]{llncs}
\mainmatter

\def\ACCV14SubNumber{70}

\usepackage[colorlinks]{hyperref}
\usepackage[usenames,dvipsnames]{xcolor}
\usepackage{xspace}
\usepackage{graphicx}
\usepackage{amsmath,amssymb} 
\usepackage[utf8]{inputenc}
\hypersetup{
	bookmarksopen=true,
	pdffitwindow=true,
	linkcolor=RoyalBlue,
	citecolor=RedViolet,
	urlcolor=RoyalBlue
}

\newcommand{\x}{\mathbf{x}}
\newcommand{\p}{\mathbf{p}}
\newcommand{\deltap}{\boldsymbol{\Delta}\p}
\newcommand{\deltax}{\boldsymbol{\Delta}\x}
\newcommand{\I}{\mathbf{I}}
\newcommand{\T}{\mathbf{T}}
\newcommand{\W}{\mathbf{W}}
\newcommand{\R}{\mathbf{R}}
\newcommand{\D}{\mathcal{D}}
\newcommand{\dWdp}{\frac{\partial \W}{\partial \p}}
\newcommand{\dTdx}{\frac{\partial T}{\partial x}}
\newcommand{\dTdy}{\frac{\partial T}{\partial y}}

\newcommand{\fig}[1]{Fig. \ref{fig:#1}}

\newcommand{\eqn}[1]{Eqn. \ref{eqn:#1}}

\newcommand{\conv}{\ast}

\newcommand{\grad}{\nabla}

\makeatletter
\DeclareRobustCommand\onedot{\futurelet\@let@token\@onedot}
\def\@onedot{\ifx\@let@token.\else.\null\fi\xspace}

\def\eg{\emph{e.g}\onedot} 
\def\ie{\emph{i.e}\onedot} 
 
\def\etc{\emph{etc}\onedot} 
\def\wrt{w.r.t\onedot} 
\def\etal{\emph{et al}\onedot}
\makeatother

\title{Regression-Based Image Alignment\\ for General Object Categories}
\titlerunning{Regression-Based Image Alignment}
\author{\href{mailto:hilton.bristow@gmail.com}{Hilton Bristow}\inst{1} and \href{mailto:slucey@cs.cmu.edu}{Simon Lucey}\inst{2}}
\authorrunning{Bristow \& Lucey}

\institute{Queensland University of Technology (QUT)\\Brisbane QLD 4000, Australia\\
\and
Carnegie Mellon University (CMU)\\Pittsburgh PA 15289, USA
}

\begin{document}
\maketitle

\begin{abstract}
Gradient-descent methods have exhibited fast and reliable performance for image alignment in the facial domain, but have largely been ignored by the broader vision community. They require the image function be smooth and (numerically) differentiable -- properties that hold for pixel-based representations obeying natural image statistics, but not for more general classes of non-linear feature transforms. We show that transforms such as Dense SIFT can be incorporated into a Lucas Kanade alignment framework by predicting descent directions via regression. This enables robust matching of instances from general object categories whilst maintaining desirable properties of Lucas Kanade such as the capacity to handle high-dimensional warp parametrizations and a fast rate of convergence. We present alignment results on a number of objects from ImageNet, and an extension of the method to unsupervised joint alignment of objects from a corpus of images.
\vspace{4mm}

\end{abstract}

\section{Introduction}
Traditionally, detectors used in general object detection have been applied in a discrete multi-scale sliding-window manner. This enables global search of the optimal warp parameters (object scale and position within the source image), at the expense of only being able to handle these simple transformations. Gradient-based approaches such as Lucas Kanade (LK)~\cite{baker_IJCV_2004}, on the other hand, can entertain more complex warp parametrizations such as rotations and changes in aspect ratio, but impose the constraint that the image function be smooth and differentiable (analytically or efficiently numerically).

This constraint is generally satisfied for pixel-based representations that follow natural image statistics~\cite{simoncelli_NEUROSCIENCE_2001}, especially on constrained domains such as faces, which are known to exhibit low-frequency gradients~\cite{cootes_report_2004}. For broader object categories that exhibit large intra-class variation and discriminative gradient information in the higher-frequencies (\ie the interaction of the object with the background) however, non-linear feature transforms that introduce tolerance to contrast and geometry are required. These transforms violate the smoothness requirement of gradient-based methods.

As a result, the huge wealth of research into gradient-based methods for facial image alignment has largely been ignored by the broader vision community. In this paper, we show that the LK objective can be modified to handle non-linear feature transforms. Specifically, we show,
\begin{itemize}
\item descent directions on feature images can be computed via linear regression to avoid any assumptions about their statistics,
\item for least-squares regression, the formulation can be interpreted as an efficient convolution operation,
\item localization results on images from ImageNet using higher-order warp parame-trizations than scale and translation,
\item an extension to unsupervised joint alignment of a corpus of images.
\end{itemize}

By showing that gradient-based methods can be applied to non-linear image transforms more generally, the huge body of research in image alignment can be leveraged for general object alignment.

\section{Image Alignment}
Image alignment is the problem of registering two images, or parts of images, so that their appearance similarity is maximized. It is a difficult problem in general, because (i) the deformation model used to parametrize the alignment can be high-dimensional, (ii) the appearance variation between instances of the object category can be large due to differences in lighting, pose, non-rigid geometry and background material, and (iii) search space is highly non-convex.

\subsection{Global Search}
For localization of general object categories, the solution has largely been to parametrize the warp by a low-dimensional set of parameters -- $x,y$-translation and scale -- and exhaustively search across the support of the image for the best set of parameters using a classifier trained to tolerate lighting variation and changes in pose and geometry. Though not usually framed in these terms, this is exactly the role of multi-scale sliding-window detection.

Higher-dimensional warps have typically not been used, due to the exponential explosion in the size of the search space. This is evident in graphical models, where it is only possible to entertain a restrictive set of higher-dimensional warps: those that are amenable to optimization by dynamic programming~\cite{felzenszwalb_PAMI_2010}. A consequence of this limitation is that sometimes underlying physical constraints cannot be well modelled:~\cite{zhu_CVPR_2012} use a tree to model parts of a face, resulting in floating branches and leaf nodes that do not respect or approximate the elastic relationship of muscles.

A related limitation of global search is the speed with which warp parametrizations can be explored. Searching over translation can be computed efficiently via convolution, however there is no equivalent operator for searching affine warps or projections onto linear subspaces.

\cite{lankinen_BMVC_2011} introduced a global method for gaining correspondence between images from general object categories -- evaluated on Pascal VOC -- based on homography consensus of local non-linear feature descriptors. They claim performance improvements over state-of-the-art congealing methods, but their only qualitative assessment is on rigid objects, so it is difficult to gauge how well their method generalizes to non-rigid object classes.

A related problem is that of co-segmentation~\cite{dai_ICCV_2013}, which aims to learn coherent segmentations across a corpus of images by exploiting similarities between the foreground and background regions in these images. Such global methods are slow, but could be used as an effective initializer for local image alignment (in the same way that face detection is almost universally used to initialize facial landmark localization).

\subsection{Local Search}
Local search methods perform alignment by taking constrained steps on the image function directly. The family of Lucas Kanade algorithms consider a first-order Taylor series approximation to the image function and locally approximate its curvature with a quadratic. Convergence to a minima follows if the Jacobian of the linearization is well-conditioned and the function is smooth and differentiable. Popular non-linear features such as Dense SIFT~\cite{liu_PAMI_2011}, HOG~\cite{dalal_triggs_CVPR_2005} and LBP~\cite{ojala_PAMI_2002} are non-differentiable image operators. Unlike pixel representations whose $\frac{1}{f}$ frequency spectra relates the domain of the optimization basin to the amount of blur introduced, these non-linear operators do not have well-understood statistical structure.

Current state-of-the art local search methods that employ non-linear features for face alignment instead use a cascade of regression functions, in a similar manner to Iterative Error Bound Minimization (IEBM)~\cite{saragih_ICPR_2006}. A common theme of these methods~\cite{kazemi_CVPR_2014,ren_CVPR_2014,xiong_CVPR_2013} is that they directly regress to positional updates. This sidesteps issues with differentiating image functions, or inverting Hessians. The drawback, however, is that they require vast amounts of training data to produce well-conditioned regressors. This approach is feasible for facial domain data that can be synthesized and trained offline in batch to produce fast runtime performance, but becomes impractical when performing alignment on arbitrary object classes, which have traditionally only had weakly labelled data.

The least squares congealing alignment algorithm~\cite{cox_ICCV_2009}, for example, has no prior knowledge of image landmarks, and learning positional update regressors for each pixel in each image is not only costly, their performance is poor when using only the surrounding image context as training data.

\cite{huang_ICCV_2007} first proposed the use of non-linear transforms (SIFT descriptors in their case) for the congealing image alignment problem, noting like us, that pixel-based representations do not work on sets of images that exhibit high contrast variance. Their entropy-based algorithm treats SIFT descriptors as stemming from a multi-modal Gaussian distribution, and clusters the regions, at each iteration finding the transform that minimizes the cluster entropy. As~\cite{cox_ICCV_2009} pointed out, however, employing entropy for congealing is problematic due to its poor optimization characteristics. As a result, the method of~\cite{huang_ICCV_2007} is slow to converge.

The related field of medical imaging has a large focus on image registration for measuring brain development, maturation and ageing, amongst others. \cite{jia_CVPR_2010,ying_NEURO_2014} present methods for improving the robustness of unsupervised alignment by embedding the dataset in a graph, with edges representing similarity of images. Registration then proceeds by minimizing the total edge length of the graph. This improves the capture of images which are far from the dataset mean, but which can be found by traversing through intermediate images. Their application domain -- brain scans -- is still highly constrained, permitting the estimation of geodesic distances between images in pixel space. Nonetheless, this type of embedding is beyond what generic congealing algorithms have achieved.

For general image categories, we instead propose to compute descent directions via appearance regression. The advantage of this approach is that the size of the regression formulation is independent of the dimensionality of the feature transform, so can be inverted with a small amount of training data.

\section{Problem Formulation}
The Inverse Compositional Lucas Kanade problem can be formulated as,
\begin{alignat}{4}
& \arg\min_{\deltap} &&\;\; || \T(\W(\x; \p)) - \I(\W(\x; \deltap)) ||^2_2
\end{alignat}
where $\T$ is the reference template image, $\I$ is the image we wish to warp to the template and $\W$ is the warp parametrization that depends on the image coordinates $\x$ and the warp parameters $\p$. This is a nonlinear least squares (NLS) problem since the image function is highly non-convex.
To solve it, the role of the template and the image is reversed and the expression is linearized by taking a first-order Taylor expansion about $\T(\W(\x; \p))$ to yield,
\begin{alignat}{4}
& \arg\min_{\deltap} && \;\; || \T(\W(\x; \p)) + \grad\T \dWdp \deltap - \I(\x) ||^2_2
\end{alignat}

$\grad \T = (\dTdx, \dTdy)$ is the gradient of the template evaluated at $\W(\x; \p)$. $\dWdp$ is the Jacobian of the template. The update $\deltap$ describes the optimal alignment of $\T$ to $\I$. The inverse of $\deltap$ is then composed with the current estimate of the parameters,
\begin{align}
\p_{k+1} = \p_{k} \circ \deltap^{-1}
\end{align}
 and applied to $\I$.

The implication is that we always linearize the expression about the template $\T$, but apply the (inverse of the) motion update $\deltap$ to the image $\I$. The consequence of this subtle detail is that $\T$ is always fixed, and thus the gradient operator $\grad \T$ only ever needs to be computed once~\cite{baker_CVPR_2001}. This property extends to our regression framework, where the potentially expensive regressor training step can also happen just once, before alignment.

For non-linear multi-channel image operators, we can replace the gradient operator $\grad \T$ with a general matrix $\R$,
\begin{alignat}{4}
& \arg\min_{\deltap} && \;\; || \T(\W(\x; \p)) + \R \dWdp \deltap - \I(\x) ||^2_2 \label{eqn:regressor}
\end{alignat}

The role of this matrix is to predict a descent direction for each pixel given context from other pixels and channels. The structure of the matrix determines the types of interactions that are exploited to compute the descent directions. If the Jacobian is constant across all iterates -- as is the case with affine transforms -- it can be pre-multiplied with the regressor so that solving each linearization involves only a single matrix multiplication.

\subsection{Fast Regression}
We now discuss a simple least squares strategy for learning $\R$. If we consider only a translational warp, the expression of \eqn{regressor} reduces to,
\begin{alignat}{4}
& \arg\min_{\deltax} && \;\; || \T(\x) + \R \deltax - \I(\x) ||^2_2 \label{eqn:translation}
\end{alignat}
where $\deltax = \deltap = (\Delta x, \Delta y)$. That is, we want to find the step size along the descent direction that minimizes the appearance difference between the template and the image. If we instead fix the $\deltax$, we can solve for the $\R$ that minimizes the appearance difference,
\begin{alignat}{4}
& \arg\min_{\R} && \;\; \sum_{\deltax \in \D} || \T(\x) + \R \deltax - \T(\x+\deltax) ||^2_2 \label{eqn:regressors}
\end{alignat}
Here we have replaced $\I(\x)$ with the template at the known displacement, $\T(\x + \deltax)$. The domain of displacements $\D$ that we draw from for training balances small-displacement accuracy and large-displacement stability. Of course, least-squares regression is not the only possible approach. One could, for example, use support vector regression (SVR) when outliers are particularly problematic with a commensurate increase in computational complexity.

Each regressor involves solving the system of equations:
\begin{alignat}{4}
& \arg\min_{\R_i} && \;\; \sum_{\deltax \in \D_i} || \T(\x_i) + \R_i \deltax - \T(\x_i+\deltax) ||^2_2
\end{alignat}
where $i$ represents the $i$-th pixel location in the image. If the same domain of displacements is used for each pixel, the solution to this objective can be computed in closed form as
\begin{alignat}{4}
& \R^*_i &=& \left( \deltax \deltax^T + \rho\I\right)^{-1} \left(\deltax^T \left[\T(\x_i+\deltax) - \T(\x_i)\right]\right)
\end{alignat}

The first thing to note is that $(\deltax^T \deltax + \rho\I)^{-1}$ is a $2 \times 2$ matrix dependent only on the domain size chosen, and not on pixel location, and can thus be inverted once and for all.

The $\deltax^T \left[\T(\x_i+\deltax) - \T(\x_i)\right]$ term within the expression is just a sum of weighted differences between a displaced pixel, and the reference pixel. \ie,
\begin{align}
\left[ \begin{array}{c}
\sum_{\Delta x} \sum_{\Delta y} \Delta x ( \T(x + \Delta x, y + \Delta y) - \T(x, y) \\
\sum_{\Delta x} \sum_{\Delta y} \Delta y ( \T(x + \Delta x, y + \Delta y) - \T(x, y) 
\end{array} \right] \label{eqn:single-regressor-exploded}
\end{align}

Other regression-based methods of alignment such as~\cite{xiong_CVPR_2013} leverage tens of thousands of warped training examples during offline batch learning to produce fast runtime performance on a single object category (faces). We cannot afford such complexity if we're going to perform regression and alignment on arbitrary object categories without a dedicated training time. 

If we sample $\deltax$ on a regular grid that coincides with pixel locations, then \eqn{single-regressor-exploded} can be cast as two filters -- one each for horizontal weights $\Delta x$, and vertical weights $\Delta y$,

\begin{align}
f_x = \left[ \begin{array}{ccc}
x_{-n} & \dots & x_n \\
\vdots &&\\
x_{-n} & \dots & x_n
\end{array} \right]
\;\;\;\;\;\;
f_y = \left[ \begin{array}{ccc}
y_{-n} & \dots & y_{-n} \\
\vdots &&\\
y_n & \dots & y_n
\end{array} \right]
\end{align}

If the $x$ and $y$ domains are both equal and odd, the contribution of $\T(x,y)$ is cancelled out. This is clearly a generalization of the central difference operator, which considers a domain of $[-1, 1]$, and forms the filters,
\begin{align}
f_x = \left[ \begin{array}{ccc}
-1 & 0 & 1
\end{array} \right]
\;\;\;\;\;\;
f_y = \left[ \begin{array}{c}
-1 \\ \phantom{+}0 \\ 1
\end{array} \right]
\end{align}

Thus, an efficient realization for learning a regressor at every pixel in the image is,
\begin{alignat}{4}
\R = (\deltax^T \deltax + \rho\I)^{-1} \left[ f_x \conv \T(\x) \;\;\; f_y \conv \T(\x) \right]
\end{alignat}
where $\conv$ is the convolution operator. For an image with $N$ pixels, $K$ channels and a warp with $P$ motion parameters, the complexity of our image alignment can be stated as a single $O(KN \log KN + KNP)$ pre-computation of the regressor, followed by an $O(KNP)$ matrix-vector multiply and image warp per iteration, with an overall linear rate of convergence.

\subsection{Regressors on Feature Images}
\begin{figure}
\includegraphics[trim=0 30 0 60,clip,width=\columnwidth]{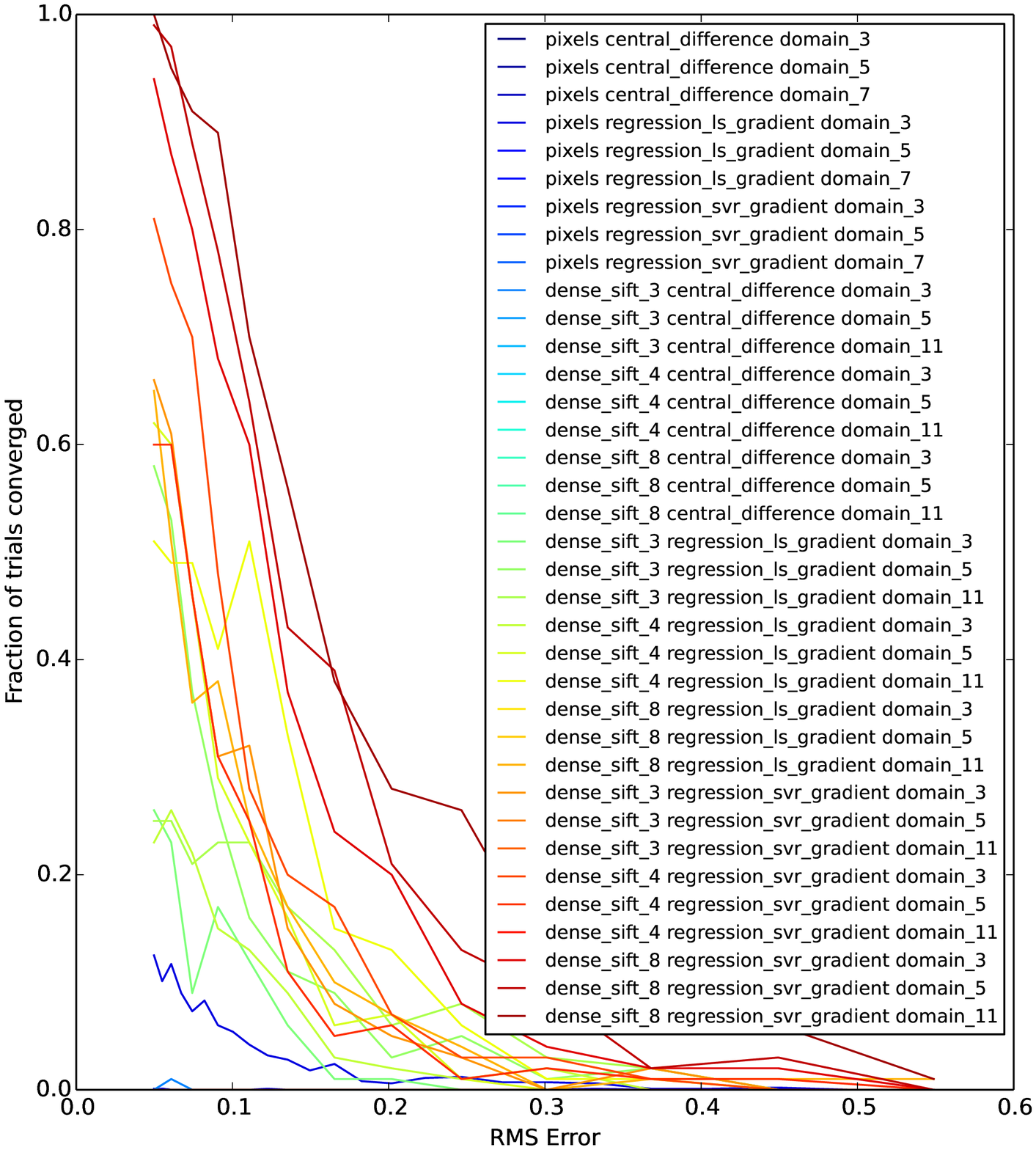}
\begin{minipage}[t]{0.5\textwidth}
\includegraphics[width=\columnwidth]{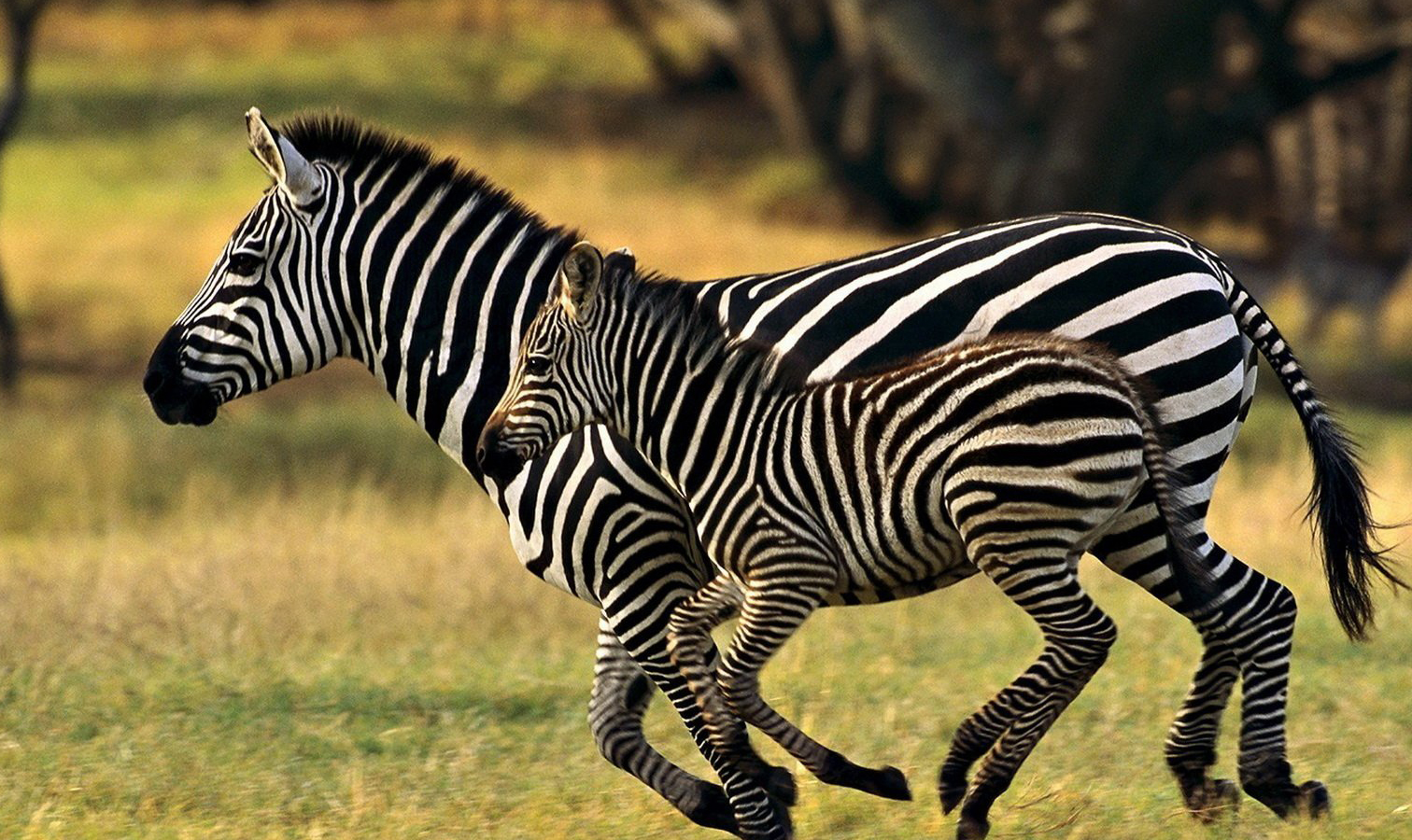}
\end{minipage}
\begin{minipage}[t]{0.5\textwidth}
\includegraphics[width=\columnwidth]{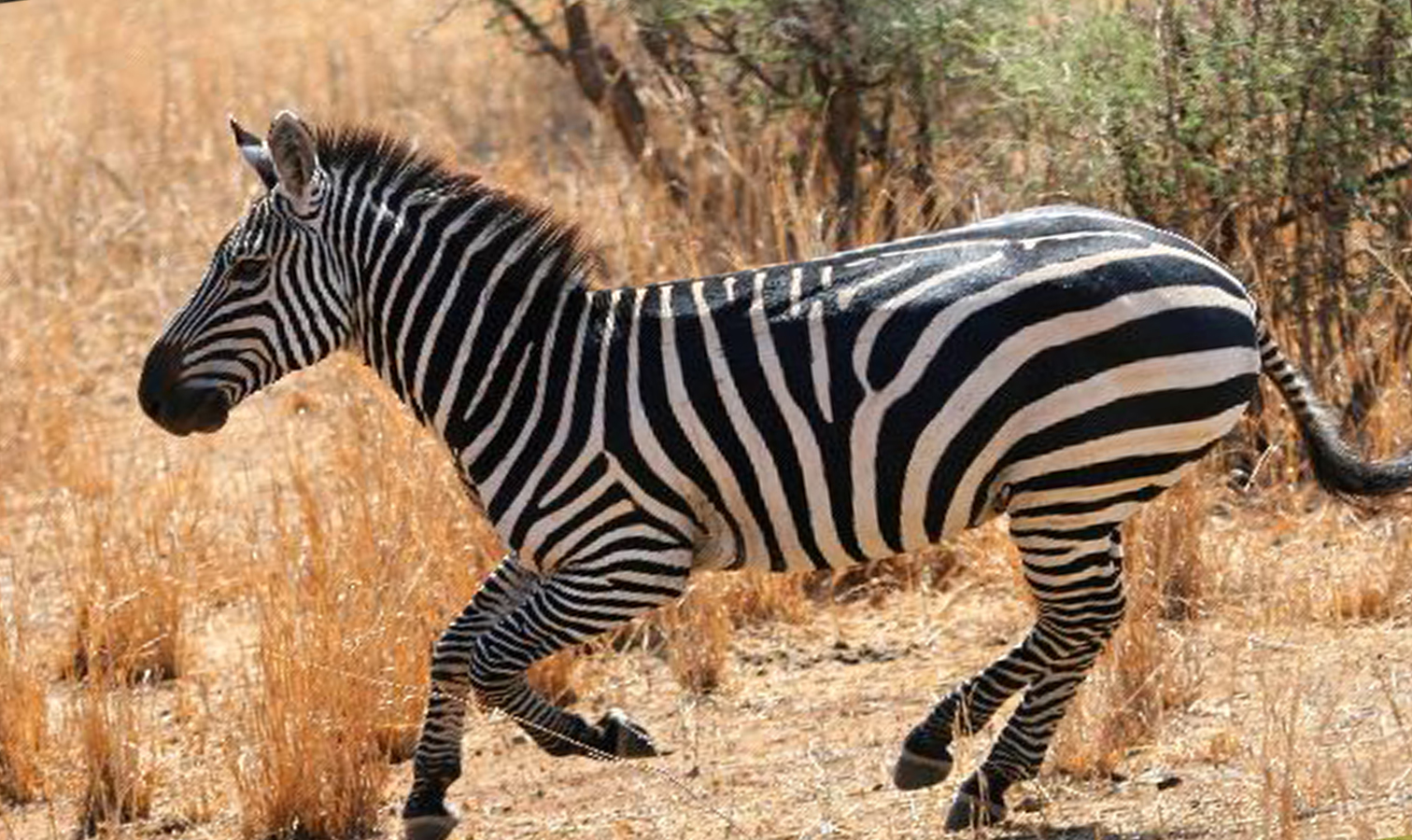}
\end{minipage}
\caption{Pairwise (LK) alignment performance of different methods for increasing initialization error. The number after Dense SIFT indicates the spatial aggregation (cell) size of each SIFT descriptor. The domain is the limit of displacement magnitude from which training examples are gathered for the regressors, or the blur kernel size in the case of central differences. There is a progressive degradation in performance from SVR to least-squares regression to central differences on Dense SIFT. The pixel-based methods fail to converge even when close to the ground truth on challenging images such as the zebra.
\label{fig:ground-truth}
}
\end{figure}

Dense non-linear feature transforms can be viewed as mapping each scalar pixel in a (grayscale) image to a vector $\R \to \R^K$. The added redundancy is required to decorrelate the various lighting transforms affecting the appearance of objects. Some feature transforms such as HOG~\cite{dalal_triggs_CVPR_2005} also introduce a degree of spatial insensitivity for matching dis-similar objects, though we find in practice that alignment performance is more sensitive to lighting than geometric effects \mbox{(\fig{ground-truth}).} 

During alignment, spatial operations are applied across each channel independently. In particular, our regression formulation does not consider correlations between channels, so separate regressors can be learned on each feature plane of the image, then concatenated. This admits a highly efficient representation in the Fourier domain -- the filters $f_x$ and $f_y$ only need to be transformed to the Fourier domain once per image, rather than once per channel.

To illustrate the benefit of applying non-linear transforms, we performed an alignment experiment between pairs of images with ground-truth registration, and progressively increased the initialization error, measuring the overall number of trials that converged back to ground-truth (within $\epsilon$ tolerance). Faces with labelled landmarks constitute a poor evaluation metric because of the proven capacity for pixel representations to perform well. Instead, we adopted the following strategy for defining ground-truth image pairs for general object classes: we manually sampled similar images from ImageNet and visually aligned them \wrt an affine warp, then ran both LK and SIFT Flow at the ``ground-truth'' and asserted they did not diverge from the initialization (refining the estimate and iterating where necessary).

For each value of the initialization error, we ran $1000$ trials. \fig{ground-truth} presents the results, with a representative pair of ground-truth images. There is a progressive degradation in performance from SVR to least-squares regression to central differences on all of the Dense SIFT trials.

Importantly, the pixel-based trials fail to converge even close to the ground-truth -- the background distractors and differences between the zebras dominate the appearance, which results in incoherent descent predictions. At the other end of the spectrum, SVR consistently outperforms least-squares regression by a large margin, indicative that a large number of sample outliers exist over both small and large domain sizes. This highlights the benefit of treating alignment as a regression problem rather than computing numeric approximations to the gradient (\ie central differences), and suggests that excellent performance can be achieved with commensurate increase in computational complexity. 

\section{Experiments}
In all of our alignment experiments, we extract densely sampled SIFT features~\cite{liu_PAMI_2011} on a regular grid with a stride of $1$ pixel. We cross-validated the spatial aggregation (cell) size, and found $4 \times 4$ regions to work best for least-squares regression, and $8 \times 8$ regions to work best for SVR. Whilst the method is certainly applicable to HOG and other feature transforms, we consider here only Dense SIFT. In the visualizations that follow, results are presented using least-squares regression. 

\begin{figure}
\includegraphics[width=\columnwidth]{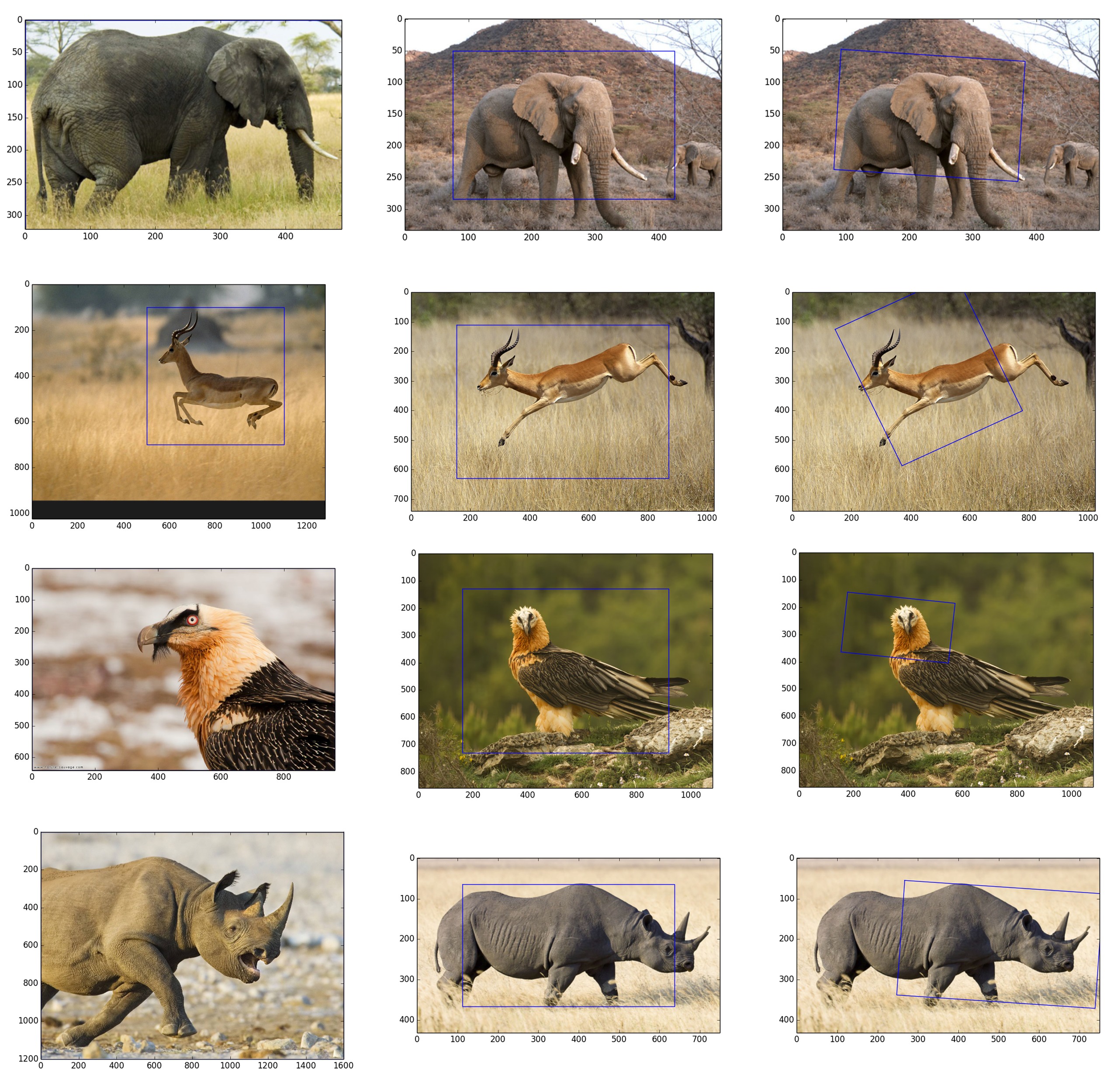}
\caption{Representative pairwise alignments. Column-wise from left to right: (i) The template region of interest. (ii) The image we wish to align to the template. The bounding box initialization covers $\approx 50\%$ of the image area, to reflect the fact that objects of interest rarely fill the entire area of an image. (iii) The predicted region that best aligns the image to the template. The four examples exhibit robustness to changes in pose, rotation, scale and translation, respectively.
\label{fig:gallery}}
\end{figure}

\subsection{Pairwise Image Alignment}
We test the performance of our algorithm on a range of animal object categories drawn from ImageNet. In \fig{gallery}, the first column is the template image. If no bounding box is shown, the whole image is used as the template. The second column shows the image we wish to align, with the initialization bounding the middle $50\%$ of pixels -- owing to the fact that photographers rarely frame the object of interest to consume the entire image area. The third column shows the converged solutions. In all of the cases shown, pixel-based representations failed to converge.

\subsection{Unsupervised Localization}
The task of unsupervised localization is to discover the bounding boxes of objects of interest in a corpus of images with only their object class labelled. In approaches such as Object Centric Pooling~\cite{russakovsky_ECCV_2012}, a detector is optimized jointly with the estimated locations of bounding boxes. Importantly, bounding box candidates are sampled in a multi-scale sliding-window manner, perhaps across a fixed number of aspect ratios. Exhaustive search cannot handle more complex search spaces, such as rotations.

Gradient-based methods derived from the Lucas Kanade algorithm such as least squares congealing~\cite{cox_ICCV_2009} and RASL~\cite{peng_PAMI_2012} have performed well on constrained domains (\eg faces, digits, building façades), but not on general object categories. Here we show that our feature regression framework can be applied to perform unsupervised localization.

The RASL algorithm performs alignment by attempting to minimize the rank of the overall stack. This only applies to linearly correlated images, however. General object categories that exhibit large appearance variation and articulated deformations are unlikely to form a low-rank basis even when aligned. The introduction of feature transforms also explodes the dimensionality of the problem, making SVD computation infeasible. Finally, RASL has a narrow basin of convergence, requiring that the misalignment can be modelled by the error term so that the low rank term is not simply an average of images in the stack (which is known to result in poor convergence properties~\cite{cox_ICCV_2009}).

We therefore present results using the least squares congealing algorithm. It scales to large numbers of feature images, shares the same favourable inverse compositional properties as Lucas Kanade, and is robust to changes in illumination via dense SIFT features.

\begin{figure}
\includegraphics[width=\textwidth]{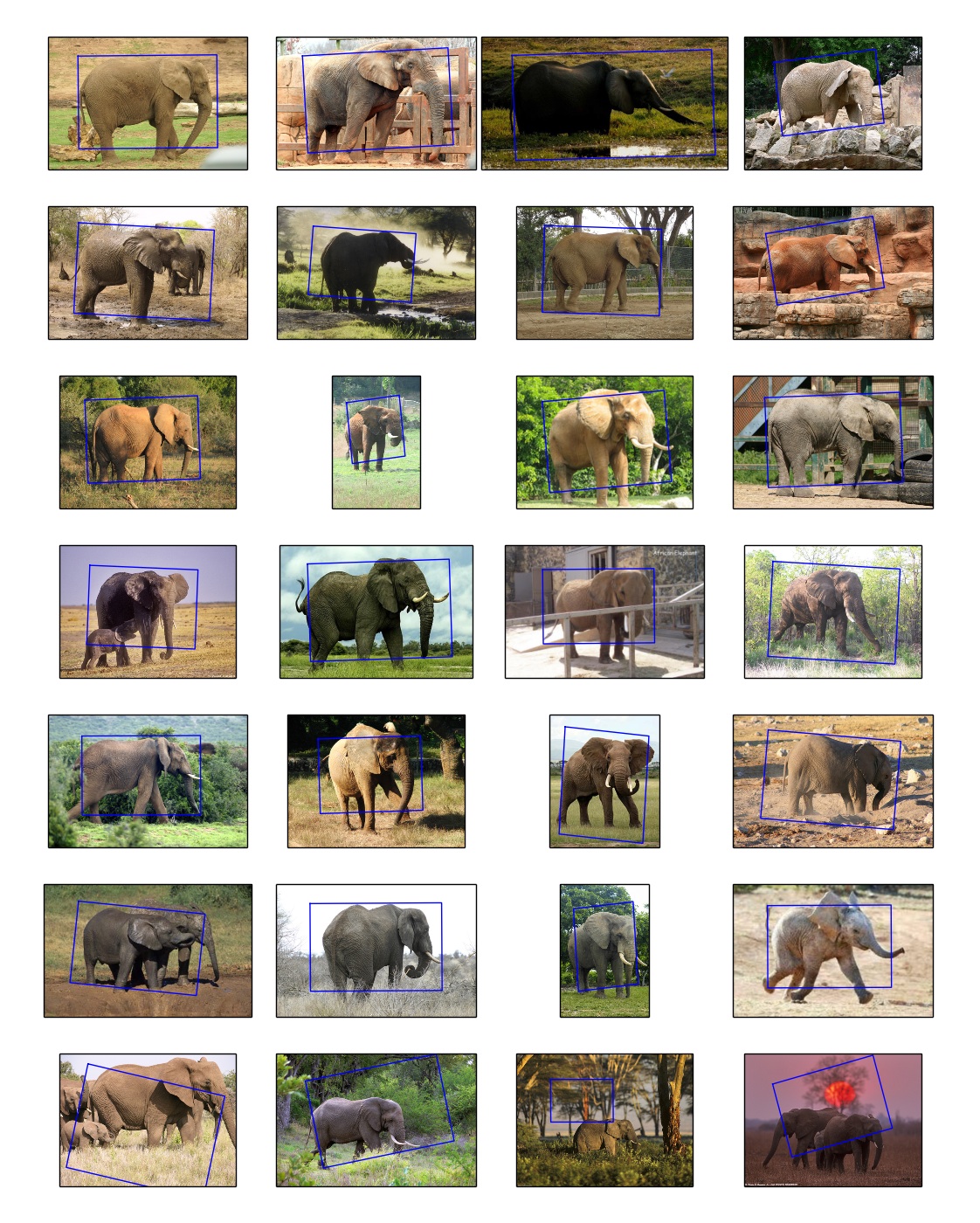}
\caption{The results of unsupervised ensemble alignment (congealing) on a set of 170 elephants taken from ImageNet. The objective is to jointly minimize the appearance difference between all of the images in a least-squares sense -- no prior information or training is required. The first 6 rows present exemplar images from the set that converged. The final row presents a number of failure cases.
\label{fig:congealing}}
\end{figure}

\begin{figure}
\begin{minipage}[t]{0.5\textwidth}
\includegraphics[width=\columnwidth]{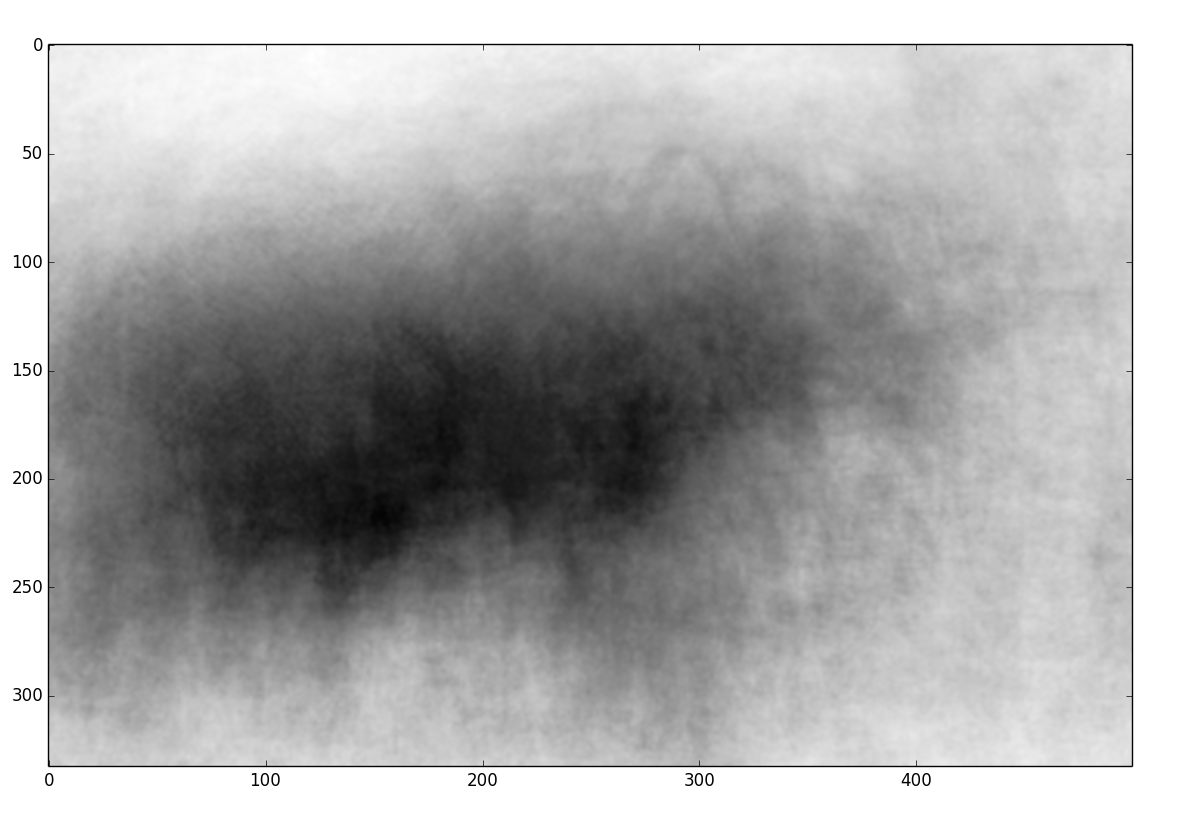}
\end{minipage}
\begin{minipage}[t]{0.5\textwidth}
\includegraphics[width=\columnwidth]{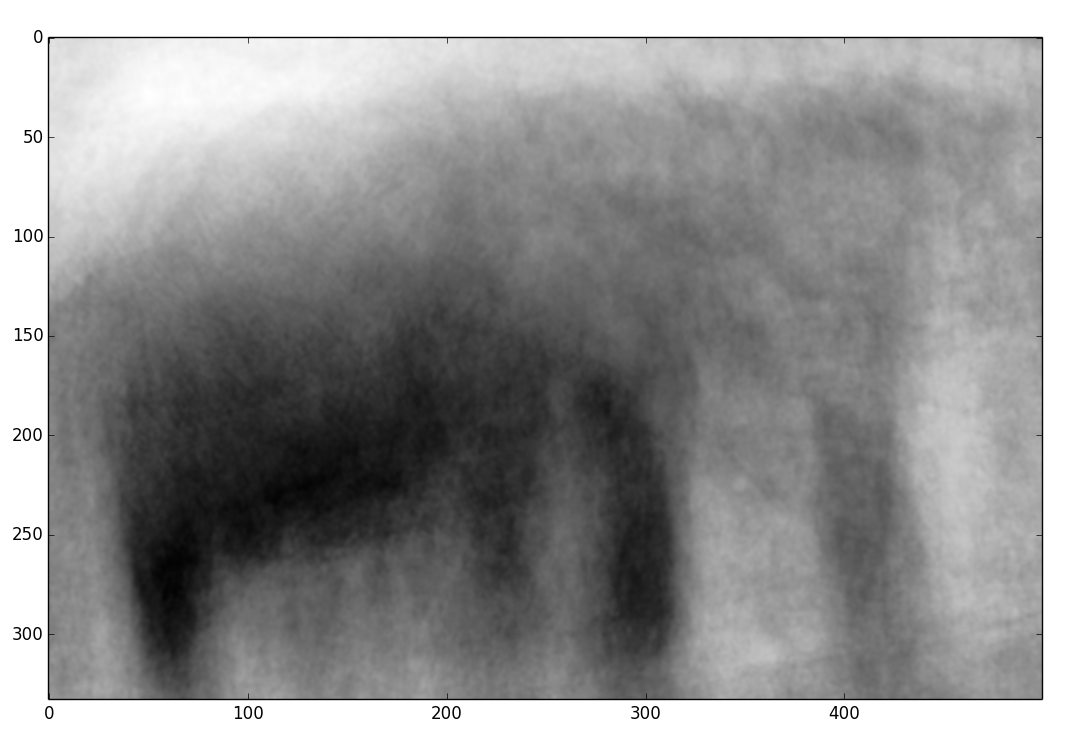}
\end{minipage}
\caption{The mean image (i) before alignment and, (ii) after alignment \wrt an affine warp. Although individual elephants undergo different non-rigid deformations, one can make out an elephant silhouette in the aligned mean.
\label{fig:mean}}
\end{figure}

\fig{congealing} shows the results of aligning a set of elephants. Recall that there is no oracle or ground truth -- the elephants are ``discovered'' merely as the region the aligns most consistently across the entire image stack. \fig{mean} illustrates the stack mean before and after congealing. Even though individual elephants appear in different poses, the aligned mean clearly elicits an elephant silhouette.

\newpage
\section{Conclusion}
Image alignment is a fundamental problem for many computer vision tasks, however a large portion of the research that has focussed on alignment in the facial domain has not generalized well to broader image categories. As a result, exhaustive search strategies have dominated general image alignment. In this paper, we showed that regression over image features could be used within a Lucas Kanade framework to robustly align instances of objects differing in pose, illumination, size and position, and presented a range of results from ImageNet categories. We also demonstrated an example of unsupervised image alignment, whereby the appearance of an elephant was automatically discovered in a large number of images. Our future work aims to parametrize more complex warps so that objects can be matched across greater pose and viewpoint variation.

\bibliographystyle{ieee}
\bibliography{citations}

\end{document}